%% file: humanoids23.tex
\documentclass[letterpaper, 10 pt, conference]{ieeeconf}  % Comment this line out if you need a4paper

\IEEEoverridecommandlockouts                              % This command is only needed if 
\usepackage{mathptmx} % assumes new font selection scheme installed
\usepackage{times} % assumes new font selection scheme installed
\usepackage{amsmath} % assumes amsmath package installed
\usepackage{amssymb}  % assumes amsmath package installed

\usepackage{graphicx}

\usepackage{colortbl}
\usepackage[normalem]{ulem}

\usepackage{mathtools}
\usepackage{cite}
\usepackage{hyperref}
\usepackage{dirtytalk}
\usepackage[inkscapearea=page]{svg}
\usepackage{adjustbox}

\newcommand{\OUT}[1]{}

\title{\LARGE \bf Robots Taking Initiative in Collaborative Object Manipulation:\\
Lessons from Physical Human-Human Interaction}
\author{Zhanibek Rysbek$^{1*}$, Ki Hwan Oh$^{1*}$, Afagh Mehri Shervedani$^{1}$, Timotej Klemen\v ci\v c$^{2}$,  Milo\v s \v Zefran$^{1}$, \\and Barbara Di Eugenio$^{3}$% <-this % stops a space
\thanks{$^{*}$ First two authors contributed equally to this work.}
\thanks{$^{1}$Z. Rysbek, K.H. Oh and Milo\v s \v Zefran are with the Robotics Lab,  Department of Electrical and Computer Engineering, the University of Illinois at Chicago, Chicago, IL 60607, USA.}%
\thanks{$^{2}$T. Klemen\v ci\v c is with the Laboratory of Robotics, Faculty of Electrical Engineering, University of Ljubljana}
\thanks{$^{3}$B. Di Eugenio is with the Natural Language Processing Lab,  Department of Computer Science, the University of Illinois at Chicago, Chicago, IL 60607, USA.}%
\thanks{ This work has been supported by the National Science Foundation grants IIS-1705058 and CMMI-1762924.}
}%

\begin{document}

\maketitle

\begin{abstract}
Physical Human-Human Interaction (pHHI) involves the use of multiple sensory modalities. Studies of communication through spoken utterances and gestures are well established, but communication through force signals is not well understood. In this paper, we focus on investigating the mechanisms employed by humans during the negotiation through force signals, and how the robot can communicate task goals, comprehend human intent, and take the lead as needed. To achieve this, we formulate a task that requires active force communication and propose a taxonomy that extends existing literature. Also, we conducted a study to observe how humans behave during collaborative manipulation tasks. An important contribution of this work is the novel features based on force-kinematic signals that demonstrate predictive power to recognize symbolic human intent. Further, we show the feasibility of developing a real-time intent classifier based on the novel features and speculate the role it plays in high-level robot controllers for physical Human-Robot Interaction (pHRI). This work provides important steps to achieve more human-like fluid interaction in physical co-manipulation tasks that are applicable and not limited to humanoid, assistive robots, and human-in-the-loop automation.

% During our preliminary data analysis, we discovered several new features that help us better understand how the interaction progresses. 

% From these features, we identified distinct patterns in the data that indicate when a participant is expressing their intent. Our study provides valuable insight into how humans collaborate physically, which can help us design robots that behave more like humans in such scenarios.
\end{abstract}

\input{Introduction.tex}

\input{Methodology.tex}

% Experiment description
% Extraction of events

% \input{Framework.tex}

\input{Results.tex}

% Explain what we have observed
% Physical interaction - primitives, analogy to dialog

% Clustering Results
% Importance of Events detected
% HMMs
% Statistical Analysis, ANOVA

\input{Conclusion.tex}

\bibliographystyle{IEEEtran}
\bibliography{RoMan_2023_arxiv}

\end{document}

%% file: Introduction.tex
\section{Introduction}

Collaborative object manipulation involves many sensory modalities, including gestures, body posture, eye gaze, and force exchanges. Humans often use different modalities to express their intent or devise an action plan. During Activities of Daily Living (ADL), partners usually plan ahead before engaging in a dyadic task. However, the plans are not always complete and the dyads sometimes need to resolve conflicts or miscommunication during the execution of the task. For example, dyads may discuss where approximately to relocate the dining table but omit details such as the final orientation of the table, or how to navigate the table around other furniture in the room. Such remaining details are typically resolved by employing one or more sensory modalities during the execution of the task. While we have tools to interpret spoken utterances and gestures~\cite{abbasi_multimodal_2019}, communication through force signals is not well-studied. But given that humans routinely communicate that way, robots need to have the capability to both understand force signals and express their intent through them if they are to engage in human-like physical interaction.

\begin{figure}[t]
    \centering
    \includegraphics[width=1.05\columnwidth, trim={6cm 4cm 7cm 4cm}, clip]{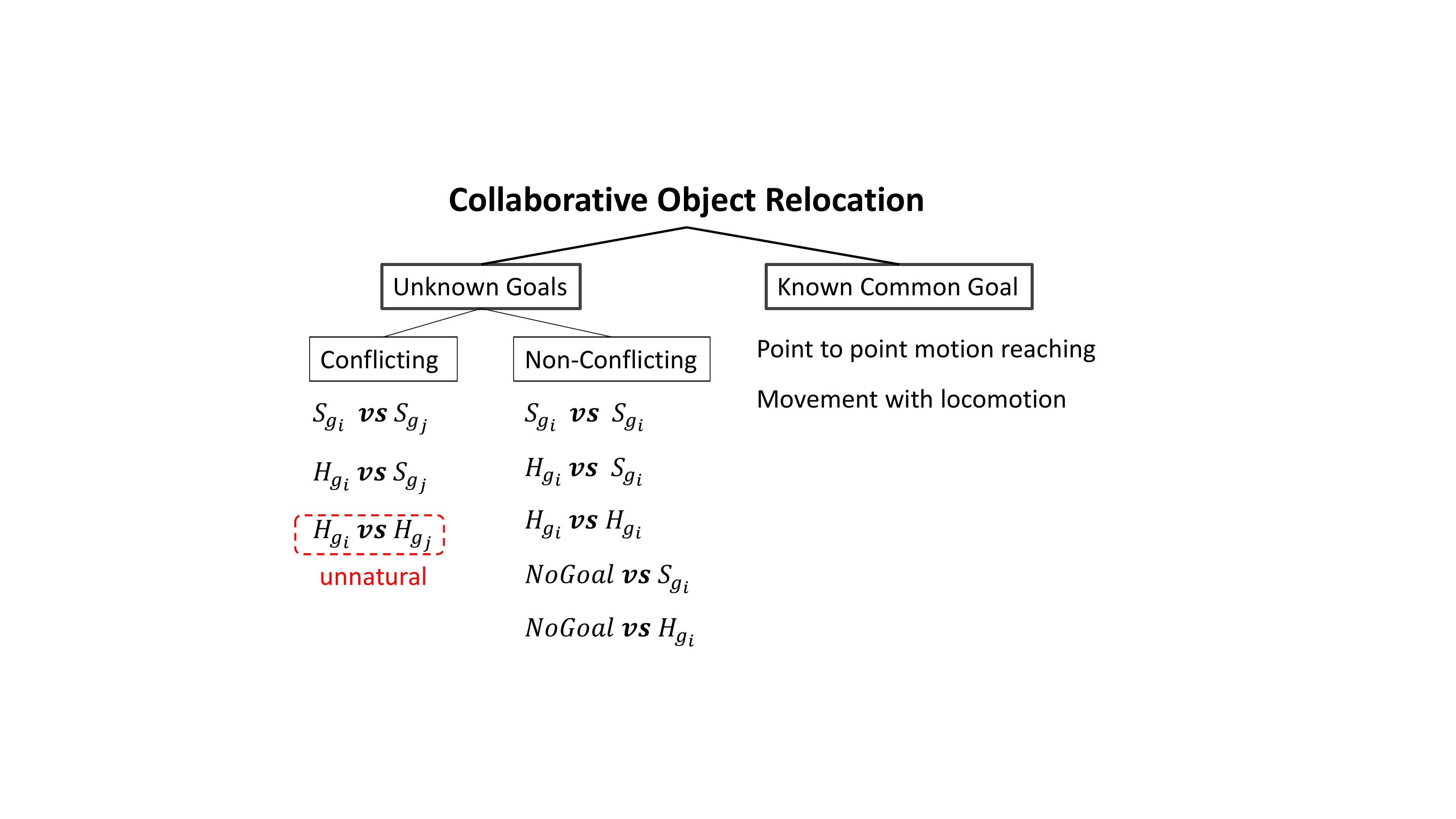}
    \caption{Goal taxonomy for collaborative object relocation. Subscript $g_i$ represents soft goal to a goal direction $g_i$ ($i \neq j $).}
    \label{fig:goal_taxonomy}
    \vspace{-0.7cm}
\end{figure}

Negotiating to reach a consensus is an integral part of successful collaboration. Quite often, literature on HRI assumes that a robot plays a subordinate role to the human in collaborative tasks. A widely used paradigm for such applications is programming by demonstration \cite{campbellProbabilisticMultimodalModeling2019, draganMovementPrimitivesOptimization2015, billard_calinon_schaal_pbd_2008} that formulates human behavior as repeatable actions. Robot response is often modulated by impedance control or admittance control \cite{hoganImpedanceControl1985, rozo_learning_2013}. On the other hand, aspects of human-human interactions are studied by haptics and kinematic metrics in virtual environments \cite{groten_haptic_dominance,mortl_role_2012, madan_recognition}. Authors in \cite{noohi_Model}, propose a model that computes interaction force for ballistic motions, where trajectory closely follows minimum jerk \cite{MinJerk_FlashHogans1985}. Human-robot interaction strategies are studied by distinguishing different haptic cues \cite{dumora_experimental_study}. In recent work, \cite{alsaadi_resolving_conflicts} proposed a pHRI system with the ability to detect conflicts based on the metrics suggested in \cite{noohi_quantitative_2014-2}. Given the scant existing work, our focus is on force communication, and in particular the mechanisms employed by the humans during the negotiation through force signals. It is only after we have a good understanding of human behavior that we can hope to design robots that can imitate us.

% We are instead interested in situations where the two partners need to play equal roles, and in particular where both are able to take initiative. For example, during the assembly, the robot might see the area of interest better so it should guide the motion. Similarly, the robot might see an obstacle that suddenly appears behind the human and should steer the motion appropriately. Of course, in general, the two participants will use multiple modalities to communicate. 

% We ground our study on a task that involves active force 

To explore force communication mechanisms that humans exploit, we ground our study on a collaborative relocation task with uncertain goals that require active communication over a haptic medium. The rest of the paper is structured as follows: in Section \ref{sec:goalTaxonomy}, we present the taxonomy that categorizes human behaviors conditioned on the available task-related information. Then, in Section \ref{sec:human_study} we describe a human study experiment during which participants negotiate a direction following each category in the taxonomy. In Section \ref{sec:Deliberate}, we propose and discuss features designed to detect and describe human intent based on force-kinematic signals. Section \ref{sec:result} presents the statistical correlation of categories in the taxonomy based on how fast dyads reach a consensus, discusses the discriminating capacity of the proposed features from the human-human data, and demonstrates the feasibility of intent classification by clustering analysis. Finally, Section \ref{sec:conclusion} summarizes the contribution of this paper with implications for future work.

%% file: Methodology.tex
\section{Dyadic Goal taxonomy}
\label{sec:goalTaxonomy}
% \subsection{Dyadic Goal taxonomy}

When dyads engage in a manipulation task, spatial awareness is a key aspect of the interaction, as they must know where to move. Considering this, we take an approach that partners always start either with a preferred goal direction in mind or with no preference at all. Without loss of generality, if there are $N$ potential goal locations, each agent's goal is $g_k \in \{ NoGoal, 1, 2, ... N \}, \enspace k=1,2$, where $k$ is the agent number. $NoGoal$ implies a follower role.

Moreover, we observed different levels of commitment to a chosen goal direction by dyads from our previous work \cite{rysbek_physical_action2021}. One partner that is more committed to its goal leads and dominates the interaction, or vice versa, a less committed partner tends to agree by compromising their intent. In daily life, such situations occur when a person who has a better view of the obstacles is likely to lead. In order to model the variations of commitment, we introduce \textit{hard} and \textit{soft} goal configuration. \textit{Hard} goal configuration requires the subject to go to the assigned goal position and they have to convince the partner to comply even if they disagree. Whereas, \textit{soft} goal configuration means that the goal would be the preferred location if no opposition is recognized from the partner. We note that this would be an example of dynamic leader-follower behavior, although it differs from what is typically studied in the literature \cite{noohi_Model, thobbi_using_2011, tsiamis_2015_cooperative}. 

\begin{figure}
    \centering
    \includegraphics[width=0.85\columnwidth, trim={6cm 2.5cm 8.8cm 1cm}, clip]{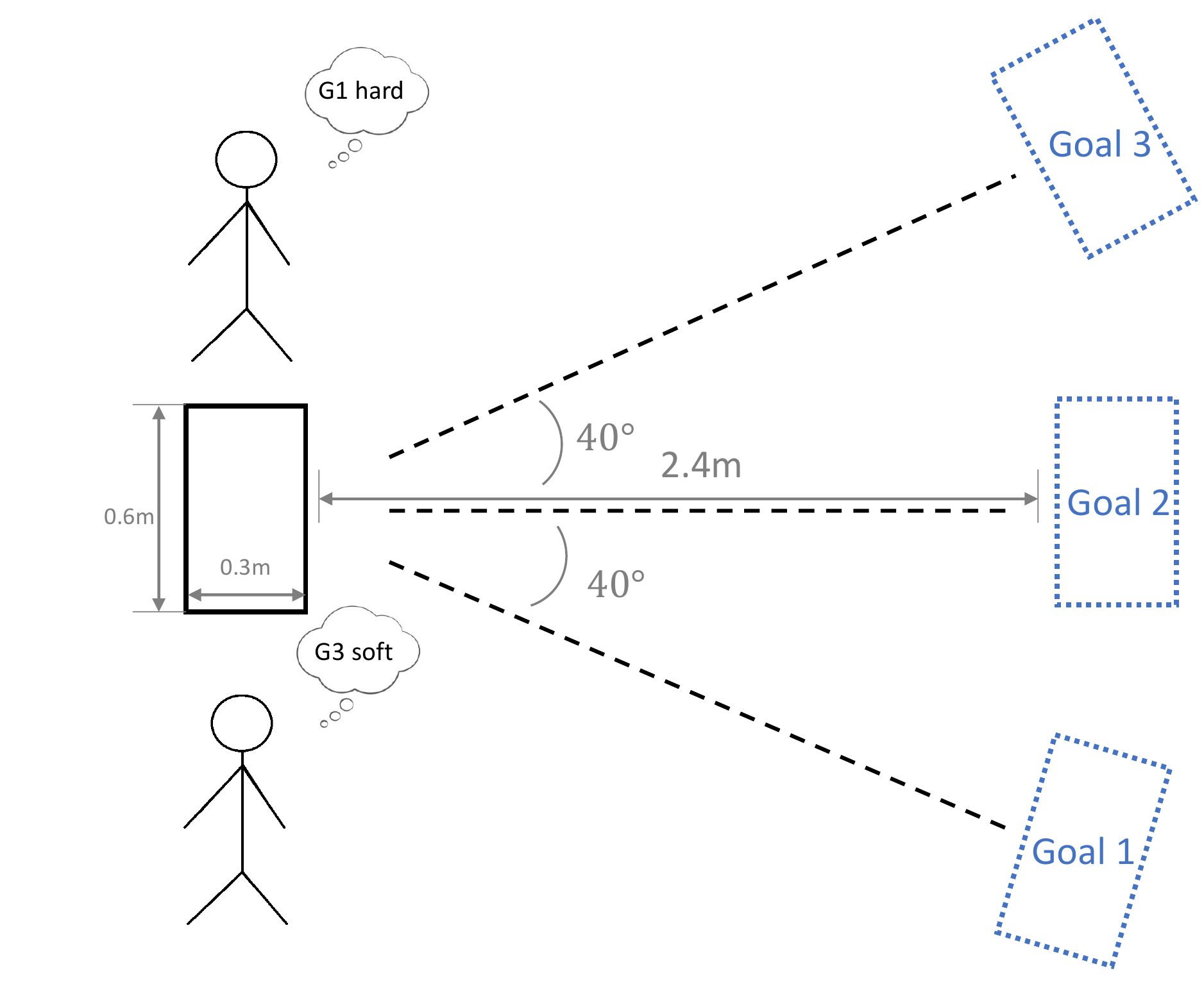}
    \caption{Experimental Environment}
    \label{fig:exp_environment}
    \vspace{-.5cm}
\end{figure}

Fig.~\ref{fig:goal_taxonomy} summarizes all possible goal combinations for the agents. In the hierarchy of the goals, we have \textbf{unknown goal} (UG) and \textbf{Known Common Goal} (KCG). KCG represents the case when partners explicitly talk about the intended goal before starting the task and the final goal configuration is agreed upon. This type of interaction does not have an explicit directional negotiation. An example of KCG cases is point-to-point motions with/without locomotion. The trajectory of the object is similar to minimum jerk trajectories \cite{noohi_Model, MinJerk_FlashHogans1985}. In UG cases, however, partners do not know each other's intended goals. Therefore, interactions can evolve into \textbf{conflicting} and \textbf{non-conflicting} scenarios. Depending on the level of commitment there could be 5 unique non-conflicting cases without repetitions (Fig.~\ref{fig:goal_taxonomy}). Unlike in KCG, dyads take extra time to perceive the intent of the partner as it is not clear if both have the same goal or not.
% \koh{KCG represents scenarios where partners engage in explicit communication about their intended goals before initiating a task. This type of interaction does not involve any directional negotiation and the final goal configuration is not concealed from either partner.} 

Conflicting interactions may unfold in 3 unique cases. In conflicting \textit{hard vs soft} goal configurations, interaction unfolds in the favor of a \textit{hard} goal dyad. However, conflicting \textit{soft vs soft} interaction is prone to be dynamic as both partners are equally committed to different goals, and interaction may take longer. In this case, force exchange may take more back-and-forth turns before the conflict is resolved. In the \textit{Hard vs Hard} case with conflicting goals, the conflict can not be resolved since both agents may refuse to compromise. We will therefore not study such interactions further.

An important motivation for this work is to explore conflicting UG scenarios that are less studied in the literature. Many of the existing studies focus on reaching motions \cite{noohi_Model} and motions with locomotion \cite{campbellProbabilisticMultimodalModeling2019} that can be classified as KCG (see Fig. \ref{fig:goal_taxonomy}). Interaction state classifier in \cite{alsaadi_resolving_conflicts} detects conflicting or non-conflicting states, which allowed the robot to switch to proactive collaboration in non-conflicting instances, and switched back to a passive follower when conflict is encountered. This interaction scenario can be thought of as a special case of the goal taxonomy where the robot starts with \textit{NoGoal} preference and the human partner has a clear goal direction in mind (\textit{NoGoal vs Soft} or \textit{NoGoal vs Hard}). We are especially interested in the \textit{Soft vs Soft} interactions, as both partners are equally committed to their goals, have no knowledge of the partner's intent, and are willing to compromise their intent based on the partner's action. At the same time, such goals assigned to subjects elicit active negotiation and provide valuable pHRI data. In the remainder of the paper, we will thus focus on the conflicting UG interactions.

% Authors in \cite{alsaadi_resolving_conflicts} propose a pHRI system that can detect four types of interaction states: harmonious translation, harmonious rotation, conflicting translation, and conflicting rotation. The ability to detect such states allows the robot to be a proactive follower, where the robot first tries to infer the human intent, if successful contributes to motion otherwise switches back to follower mode. 

\begin{figure}
    \centering
    \includegraphics[width=\columnwidth, trim={8.5cm, 5.5cm, 11cm, 4cm}, clip]{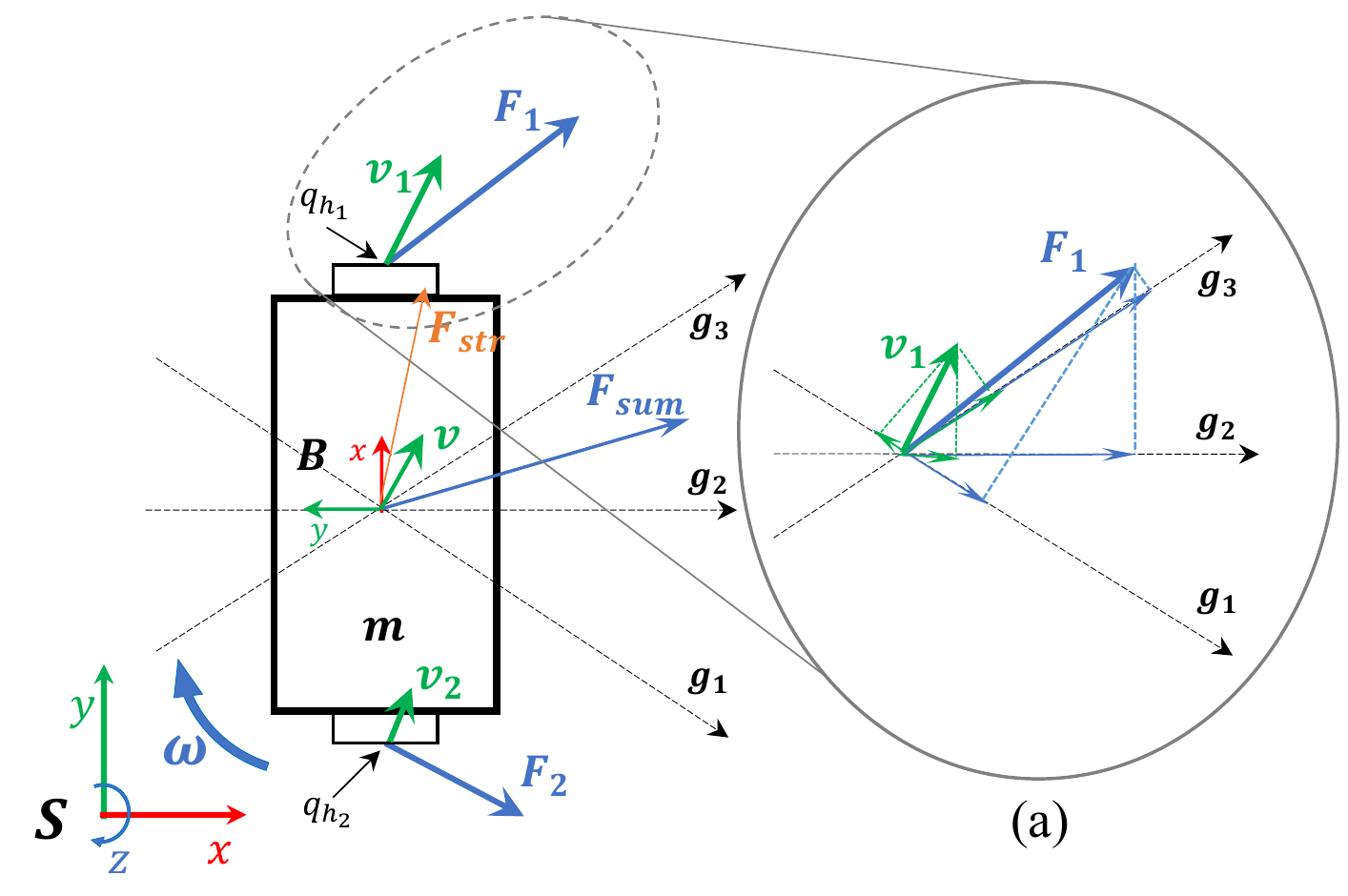}
    \caption{Free body diagram of the manipulated object in a relocation task. \textit{\textbf{S}} - ground fixed spatial frame, \textit{\textbf{B}} - body affixed frame, \textbf{\textit{m}} - object mass, $v$ - velocity of the center of mass, $v_1, v_2$ - velocities of handle 1 and 2 respectively, $F_1, F_2$ - applied forces, $\omega$ - object rotational velocity, $g_1, g_2, g_3$ - goal directions, $q_{h_1}, q_{h_2}$ are grasping points. (a) Example of the projection of force and velocity to goal directions.}
    \label{fig:free_body}
    \vspace{-.5cm}
\end{figure}

\section{Experiment}
\label{sec:human_study}

% % thoughts that are not added
% The evolution of interaction is dependent on the goals of the partner.

% This experiment is the stereotypical scenario of possible major decision-making situations that might happen in pHHI. And focus on this to get insight into a robot controller for pHRI that can take initiative.

% Symbolism in phhi comes from goal locations.

% Distinguish how much information has prior information.
% Partial information interactions

% \textcolor{red}{Motivate that this models stereotypical directional conflict during physical interaction }

% \begin{enumerate}
%     % \item Here talk about different goals. Hierarchy of goal assignment. Assumption of having a certain goal.
%     \item Observation of how humans behave. For example, we think during interaction in the kinamatic sense. And communicate over force medium to express intent.
%     \item Force decomposition.
%     \item Torques. Talk about its effect. If there is show what is the effect. If none say why.
%     \item Indices (Power, projection to goal). Elaborate each of them, and discuss the usefulness.  - Moved to next section 
% \end{enumerate}

\subsection{Task}

An experiment was designed to study directional decision-making in physical human-human interaction. A diagram of the environment is shown in Fig. \ref{fig:exp_environment}. The task is to collaboratively relocate the tray to one of three goal locations ($N=3$). Goal locations are spread out in a circular manner, where the radius is approximately 2.4m with $40^\circ$ angular spacing. Before the start, each subject privately receives randomly assigned goal configuration $g_k$ where $g_k \in \{ NoGoal, S_i, H_i \},\enspace k=1,2, \enspace i=1,2,3$. Subjects were instructed to only communicate through the haptic sense, and they are prohibited to chat, use gestures, or make eye contact. Moreover, they were asked to grasp firmly with the comfortable hand that keeps the body open during the maneuver. Two beep sounds are generated to control the initial movements. The first beep signals the time to grasp the tray, and the second beep permits them to move in the assigned direction. This allows us to measure the grasping force when they are static between the beeps and the exact initial time of the negotiation phase. Subjects engage in negotiations either to convince them to follow their goal or to give up their assigned goal. After delivering the object to the goal location, they follow the same procedure to come back to the initial position. However, there is no explicit directional decision since they both clearly know the final destination. Therefore, it can be labeled as a KCG goal from Fig.~\ref{fig:goal_taxonomy}.

The experiment was conducted across 4 days, and a total of 16 subjects were recruited from the University of Illinois at Chicago campus. In a single day, 4 participants formed 6 dyads, summing 24 dyads across 4 days. Before engaging in the task, they were given time to practice adapting to the beeps and learn the goal types. This alleviates the learning effect. The human study experiment was conducted according to IRB protocol.

\subsection{Setup}
A wooden tray is used as a manipulated object. To track the interaction we have attached two RFT60 force-torque sensors  between the handle and the tray (Robotous Inc.). 6-DOF IMU sensor was used to track the linear acceleration and the angular velocity. Both, IMU and force-torque sensors are connected to an onboard Raspberry Pi that wirelessly transmitted the data to the main computer in real-time. The sampling rate for the force-torque sensor was 200Hz, and for the IMU was 100Hz. Including onboard equipment weight of the tray was $2.3$kg and the dimensions were $61cm \times 31cm$. The 6D pose of the tray was tracked using the ArUco library from OpenCV \cite{ArUco}. Several fiducial markers were attached to the surface and the sides of the tray. To deal with occlusions, three cameras overlooking the experimental environment recorded the image streams and transmitted them to the main computer with a USB cable. Robot Operating System (ROS) is used as a data collection environment.

Extended Kalman Filters \cite{imufilter_kalman} are used to fuse angular velocity and orientations from three cameras. Subsequently, position and velocity signals were obtained using linear Kalman Filters by fusing linear acceleration and position data. All signals including force-torque, pose, twist, and accelerations were pre-processed using a low-pass filter with a cutoff frequency of $12.5$Hz.

% Several sensors were used to capture the interaction between humans during the collaborative task. The force data were collected by two RFT60 force-torque sensors (Robotous Inc.), each installed between the handle and the tray  sampled at 200Hz. The weight of the tray was $2.3 kg$, and the dimensions are $61 cm \times 31 cm$. At the center of the tray, a Raspberry Pi with an embedded 6-DOF IMU sensor (MPU9255) was attached to track the configuration (position and orientation) of the tray. To track the position of the tray, multiple fiducial (ArUco) markers are affixed to the surface and the sides of the tray, and an online position tracking algorithm was implemented using the ArUco library \cite{ArUco}. Three USB cameras were placed in a triangular configuration to record the movement of the subjects and the tray, allowing us to deal with the occlusions of the markers. The data collection was implemented in the Robot Operating System (ROS) environment. 

\begin{figure}
    \centering
    \includegraphics[width=\columnwidth, trim={9.5cm 9cm 10cm 5cm}, clip]{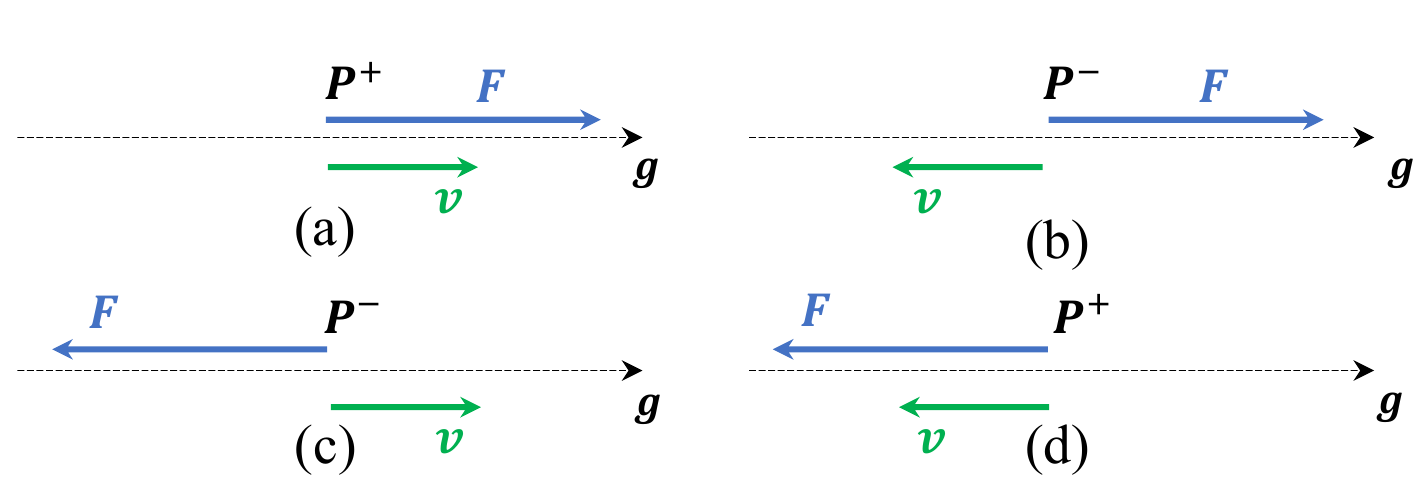}
    \caption{Combination of the signs of force and velocity and resulting power.}
    \label{fig:fv_dir}
    \vspace{-.5cm}
\end{figure}

\begin{figure*}[t]
    \centering
    % \adjustbox{trim=5.5cm 0cm 5cm 0cm}{\includesvg[inkscapelatex=false, width=1.2\textwidth]{figures/power_goal_types.svg}}
    \includegraphics[width=1.\textwidth, trim={2cm 0cm 2cm 0cm}, clip]{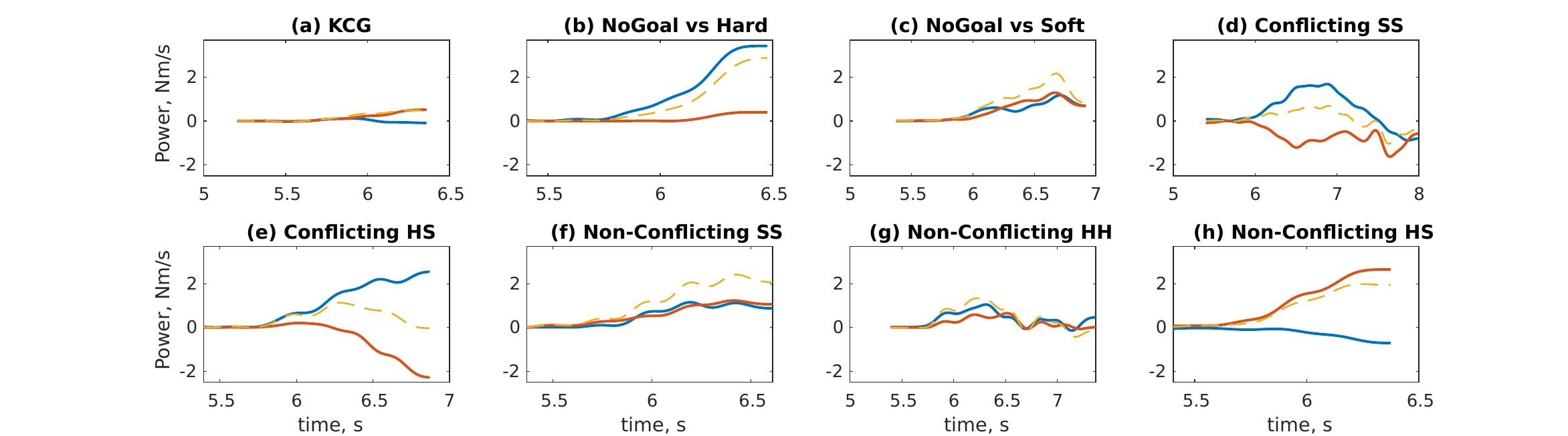}
    % \centering{\includesvg[inkscapelatex=false, width=1.05\textwidth]{figures/power_goal_types.svg}}
    \caption{Examples of power index from experimental observations by interaction types. Respective interaction types are shown on the title of the subplots. Power by subject 1 and subject 2 are denoted with blue and red curves respectively. The yellow dashed curve represents the sum of the power applied to the center of the mass.}
    \label{fig:power_by_interaction_types}
\end{figure*}

\section{Deliberate Action Metrics}
\label{sec:Deliberate}

\subsection{Force communication}

It is widely accepted that dyads communicate with force exchanges in haptically-coupled tasks. The consequence of two agents trying to move the same object results in exerting excessive effort more than what it takes to move the object. That is, $|F_{sum}| \leq |F_1| + |F_2|$,
where $F_1$, $F_2$ are forces each agent applied and $F_{sum}=m\ddot{x}=F_1+F_2$. 

The interaction force is the excessive component of applied force that does not contribute to the motion. Knowing the precise interaction force can tell us about the accurate state of the interaction. However, existing approaches that compute this force cannot be used for deliberate locomotions \cite{groten_haptic_dominance, noohi_Model, rahman_ikeura_2000}. In our task, we are interested in finding the human intent during general walking motions. This adds many confounding components of interaction force such as varying grasping force and oscillation due to walking. This increases the signal-to-noise ratio and makes it difficult to estimate the clear intent of the dyads. While grasping force can be estimated at the rest moments such as the start and the end instances of the interaction, it is not feasible to estimate it during the motion \cite{noohi_quantitative_2014-2}. Therefore, additional kinematic signals are required to interpret the human intent, unlike the haptic-only approach in \cite{alsaadi_resolving_conflicts}.

% The forces from each subject applied to each handle of the tray are depicted as $F_{1}$ and $F_{2}$ respectively. In our previous paper~\cite{rysbek_physical_action2021}, we described the sum of the forces, $F_{sum} = F_{1} + F_{2}$, directly affects the motion of the tray. Furthermore, we exploited the stretching force, $F_{stretch} = F_1 - F_2$, which we used as an indicator to determine the state of the interactions since it amplified the components of the applied forces and cancels out the forces used for motion. 

% definition of interaction force in some works regard is as a disturbance and something to avoid during interaciton. But, we regard it as a useful signal to interpret the state of conflict..

% Grasping force is a confounding factor. We can estimate very well at the beginning. We distinguish high Fgrasp from rest.
% F applied is useful when Fgrasp is high.

\subsection{Interaction Power}

Dyads use push and pull actions creating temporal conflict to express their intent. This causes a sharp rise in the magnitude of the difference of the forces ($F_1-F_2$) \cite{rysbek_physical_action2021}. The attitude toward this phenomenon varies in the literature. For example, authors in \cite{groten_haptic_dominance} and \cite{mortlRoleRolesPhysical2012} regard avoiding the conflict as merely an objective of the dyads, and the force of the agent who is not contributing to the motion is considered as the interaction force. In contrast, based on human experiments, we view temporal conflict as the consequence of symbolic actions in a conflicting goal direction. Hence, the key to interpreting agent intent is to look at the periods of such actions. An important observation is push-pull actions result in slight movement in the object. Thus, we propose to refer to the instantaneous power of each agent:
\begin{equation}
    P_{k}(t) = F_{k}\cdot v_{k} =  \|F_{k}\|\|v_{k}\|\cos(\angle(F_{k},v_{k})) \quad k=1,2
    \label{eq:power}
\end{equation}
where $k$ denotes the agent number. The velocity of the grasping point $v_k$ is measured as follows. 
\begin{equation}
    v^{s}_k = R_{sb}\cdot(\omega_{b}\times q^{b}_{k}) + \dot{p}_{sb}, \quad k=1,2
\end{equation}
where $v^{s}_k$ denotes the velocity of handle $k$ in the spatial frame (Fig. \ref{fig:free_body}), $q^{b}_{k}$ and $\omega_b$ are the handle position and the angular velocity expressed in body frame $\textbf{B}$ (center of the tray), $R_{sb}$ and $\dot{p}_{sb}$ are the rotation matrix and the velocity of frame $\textbf{B}$ relative to spatial frame $\textbf{S}$ (global reference frame) respectively. Analogously, total exerted power on the object can be computed as:
\begin{equation}
    P_{sum}(t) = F_{sum}\cdot v =  \|F_{sum}\|\|v\|\cos(\angle(F_{sum},v))
    \label{eq:power_sum}
\end{equation}
where $v$ is the velocity of frame $\textbf{B}$. Similar metrics exist in the literature, where authors in \cite{madan_recognition, feth_performance_related} use power in a scalar form to distinguish interaction patterns in the virtual environment.

Note that, positive power implies that the agent is applying force in the direction of velocity which  contributes to the motion significantly. Negative power implies that the person is resisting the motion. In the context of interaction, this could be accepted as opposition. Examples from the experiment are discussed in Section \ref{sec:result}.

% Similar to force decomposition $|P_{sum}| \leq |P_1| + |P_2|$ holds. 

% It is different from physical power. This is interaction power.

% \subsection{Handle Velocities}

% While the torques were very small compared to the forces, angular velocities were comparable with linear velocities. Moreover, we need to analyze the motions of each handle separately to fully understand the actions of each agent. Since the tray is rigid, the relative location of the handles from the center of the tray is fixed. From the online tray position tracking, we keep track of the transformation between the spatial frame $S$ (global reference frame) and the body frame $B$ (frame of the center of the tray).
% \begin{equation}
%     g_{sb} = 
%     \begin{bmatrix}
%         R_{sb} & p_{sb} \\
%         0 & 1
%     \end{bmatrix}
% \end{equation}
% We can extract the angular velocity of the tray ($\omega_{b}$) from the IMU's gyroscope readings and set the location of the handles based on the body frame as $q^{b}_{h_1}$ and $q^{b}_{h_2}$. Integrating the sensor readings, we can get the following velocities for each handle based on the spatial frame:
% \begin{equation}
%     v^{s}_{h_k} = R_{sb}\cdot(\omega_{b}\times q^{b}_{h_k}) + \dot{p}_{sb}, \quad k=1,2
% \end{equation}
% With this measurement, we were able to track the rotational actions of each subject respectively.

\subsection{Power to Goal directions}

The interaction power is suitable to determine the general direction of motion. However, the symbolism of the action is tied to the valid goal direction. One such way to interpret agent action is to look at the projection of the power in goal directions. Hence, exerted power by an agent reflected in a goal direction $g_i$ can be defined as:
\begin{equation}
\begin{split}
    P^{g_i}_{k} &= F^{g_i}_{k}\cdot v^{g_i}_{k} \\
    &= \|F_{k}\|\|v_{k}\|\cos(\angle(F_{k},g_i))\cos(\angle(v_{k},g_i))% \quad k=1,2
\end{split}
\end{equation}
And projected force and velocity for each agent $k$ are defined as follows:
\begin{align}
    F^{g_i}_{k} = \|F_{k}\|\cos(\angle(F_{k},g_i)) \\
    v^{g_i}_{k} = \|v_{k}\|\cos(\angle(v_{k},g_i)) 
\end{align}

Note that both $F^{g_i}_{k}$ and $v^{g_i}_{k}$ are scalar quantities. When an agent applies force in a particular direction, the respective projection of the force will resonate (see Fig. \ref{fig:free_body}a to goal 1). Likewise, opposition to the motion is reflected as a negative force in a goal direction (Fig. \ref{fig:free_body}a to goal 3). Power represents the manifestation of the action in a direction. Because, when there is a movement -- power is non-zero. It is important to note, that positive projected power does not always represent a positive effort to a goal (see Fig. \ref{fig:fv_dir}) because negative force and velocity will result in the positive power as shown in Fig. \ref{fig:fv_dir}d. Although, in the latter case it symbolically means moving away from that goal.

% discuss figure \ref{fig:fv_dir}

% There are four possible interactions regarding the direction of the goal from the current position of the tray as in Fig.~\ref{fig:fv_dir}. 
% The alignment of the direction of the force with a certain goal direction describes whether the subject's intent matches the current goal or not. 
% The alignment of the current motion of the tray (velocity) with a certain goal direction describes whether the object is moving toward the goal or not. 
% If the force and velocity have a common direction, then the subject tries to follow the motion of the object.
% When both are in a different direction, then the subject opposes the current motion of the tray.

%% file: Results.tex
\section{Results and Discussion}
\label{sec:result}

\begin{figure}
    \centering
    \includegraphics[width=0.9\columnwidth, trim={3cm 8.5cm 3cm 8cm}, clip]{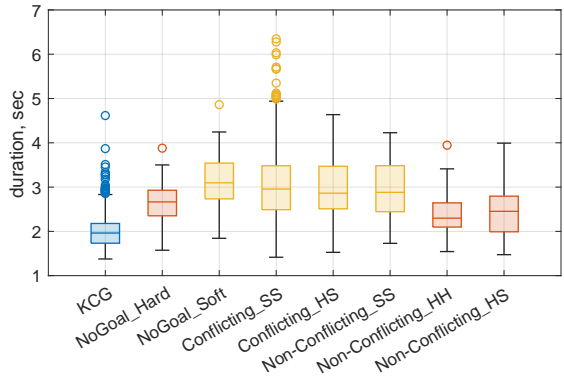}
    \caption{Box chart of negotiation phase duration by 8 different goal assignment combinations. Points outside of whiskers are outliers. Box represents the 25th and 75th percentile. Different colors represent grouping by statistical similarity.}
    \label{fig:negotiation_time}
\end{figure}

\subsection{Negotiation phase and Goal Taxonomy}
The decision-making phase in a task happens at the beginning of the motion. And it ends when agents settle into a common direction after the active negotiation phase\cite{rysbek_physical_action2021}. Force exchanges applied beyond the negotiation phase are irrelevant to directional decision-making and are detrimental in identifying actions related to major decisions. Hence, it is crucial to find the precise boundary of the negotiation phase for meaningful analysis. Beep times give us the exact moment when subjects initiate the motion, but the end time of the negotiation phase must be estimated from the data. By definition, the negotiation phase ends when subjects resolve the conflict. This implies that the direction of the object's velocity is heading toward the goal during the execution phase. One way to estimate that is as follows:
\begin{equation}
    t_{dec}: \enspace dist(p(t)+ \lambda v(t), l(g))<\epsilon, \enspace \forall t \geq t_{dec}, \enspace \lambda \in \mathbb{R} \\
\end{equation}
where $p(t)+ \lambda v(t)$ is the line defined by the object coordinate $p(t)$ and velocity $v(t)$, $l(g)$ is the coordinate of goal $g$, and $dist$ is function between the line and a point. Threshold $\epsilon$ can be different depending on the scale of the environment. In our experiment, $\epsilon=0.2m$ was chosen and the negotiation time is measured as $t_{dec}-t_{start}$. 

% In addition, the negotiation phase lasts much shorter than the total interaction time, and forces during the execution phase might overwhelm interaction forces from the negotiation. 

Fig. \ref{fig:negotiation_time} illustrates the box chart of the negotiation time from each interaction type described in Section \ref{sec:goalTaxonomy}. According to ANOVA results, KCG dyads spent the least amount of time for negotiation compared to all the other interactions ($p < 0.038$). This is expected since in UG cases subjects take extra time to perceive their partner's intent and increase the time to settle into a common goal. Negotiation time for \textit{NoGoal} vs \textit{Hard}, \textit{Non-Conflicting HH}, and \textit{Non-Conflicting HS} were significantly different from \textit{NoGoal} vs \textit{Soft} and \textit{Conflicting SS} ($p < 0.005$). And during \textit{Conflicting HS} dyads negotiated longer compared to \textit{Non-Conflicting HH} and \textit{Non-Conflicting HS} ($p < 0.003$). Tukey's honestly significant difference test was used to estimate pairwise differences. One can note that non-conflicting interactions involving one \textit{Hard} goal subject (red boxes) have a shorter mean value. This can be explained by the presence of a decisive agent that simplifies the decision-making process. Whereas, conflicting interactions (yellow boxes in Fig. \ref{fig:negotiation_time}) have larger mean values compared to the rest. Since both partners are equally committed to their goal, interaction might take a few more turns compared to the \textit{Hard} goal case.

\begin{figure}[t]
    \centering
    \includegraphics[width=\columnwidth]{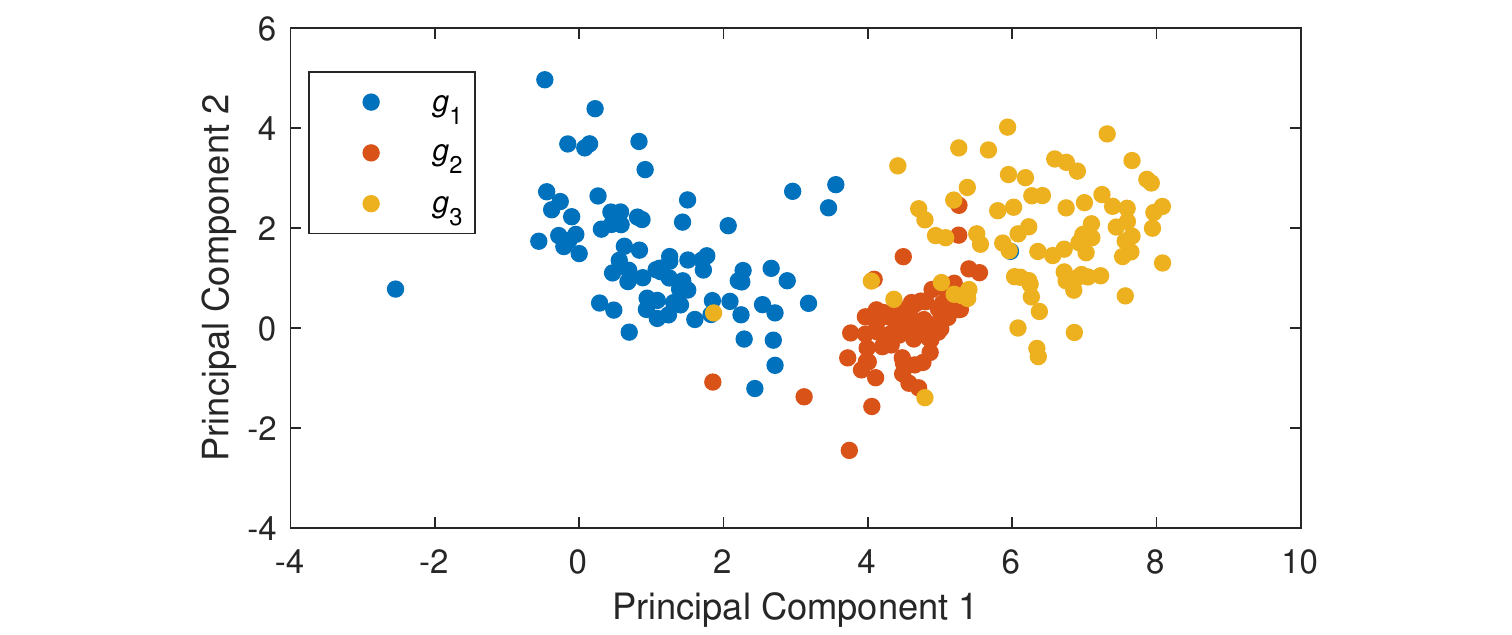}
    \caption{Strength of intent projected to 3 goals in all interactions where an agent had goal H3. The data is well clustered after LDA dimensionality reduction to two principal components.}
    \label{fig:lda}
\end{figure}

\begin{table}[t]
\caption{Clustering  scores}\label{tbl:lda_eval}
\begin{tabular}{|c|c|c|c|}
    \hline
    Assigned Goal & Calinski-Harabasz & Davies–Bouldin & Silhouette\\
    \hline
    H1 & $289.5661$ & $0.9207$ & $0.5187$\\
    H2 & $195.1549$ & $0.6994$ & $0.5650$\\
    H3 & $309.7367$ & $0.7544$ & $0.5682$\\
    \hline
\end{tabular}
\vspace{-0.5cm}
\end{table}

% \textcolor{red}{ Difference of intent between partners}

Fig. \ref{fig:power_by_interaction_types} shows power indices measured during the negotiation phase for each type of interaction. During the KCG case, dyads do not negotiate major directions. Consequently, the magnitude of applied forces and power remains low compared to UG cases. Fig. \ref{fig:power_by_interaction_types}a shows a KCG case where both subjects moved simultaneously to the goal. In some cases, KCG might resemble conflicting interactions due to delayed response. But, common traits are fast acceleration and sustained velocity of the object in the goal direction. 

% negotiate only implicit aspects of motion such as the speed, height of the grasp, and timing of the motion, and do not negotiate major directions.

Non-conflicting interactions share similar patterns where both subjects cooperate to the same motion and $P_{sum} = |P_1| + |P_2|$ holds true (Fig. \ref{fig:power_by_interaction_types}f,g,h). However, the level of commitment might change the picture. For example, in Fig. \ref{fig:power_by_interaction_types}h a subject assigned to \textit{Hard} goal (red) moved more aggressively leaving the partner behind, although \textit{Soft} goal subject (blue) did not show clear opposition.

Follower mode situations are shown in Fig. \ref{fig:power_by_interaction_types}b and c respectively against \textit{Hard} and \textit{Soft} (leader) goals. Note that, the power of the follower agent in the former case is around zero and \textit{hard} goal dominates the motion. In contrast, \textit{soft} goal case both of them similarly exert effort, which indicates the proactive behavior of the follower.

\textit{Conflicting HS} in Fig. \ref{fig:power_by_interaction_types}e shows the situation where each partner with a different intention actively moves in the opposing direction causing significant conflict. Typically, bifurcation is observed from power plots where directional disagreement (explicit negotiation) occurs. Then, the subject with the \textit{hard} goal assignment (blue) drags the object with extra force, and the object travels in its direction. 

\begin{figure*}[t]
    \centering
    % \includegraphics[width=\textwidth, trim={8cm 0.5cm 8cm 0.5cm}, clip]{figures/case_study_power_projections.png}
    % \adjustbox{trim=7cm 0cm 6.5cm 0cm}{\includesvg[inkscapelatex=false, width=1.2\textwidth]{figures/case_study_power_projections_v2.svg}}
    \includegraphics[width=1.\textwidth, trim={3cm 0cm 2cm 0cm}, clip]{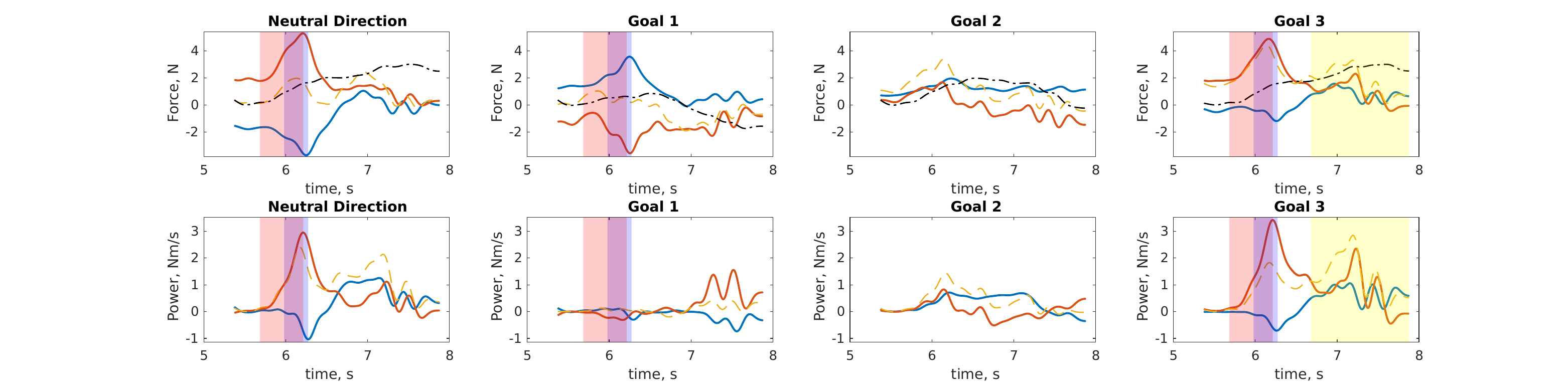}
    \caption{Force, velocity, and power and their projection to goal directions. Force and velocity are shown in the first row, and the power is in the second. Subject 1 and subject 2 are denoted in blue and red curves respectively, the yellow dashed curve is the sum of the force (first row) and power (second row). The black dashed curve is scaled velocity. Blue and red shaded areas represent the action periods for subjects 1 and 2 respectively. The yellow-shaded area is the execution phase. In this example, subject 1 was assigned to $S_1$ and subject 2 to $S_3$.}
    \label{fig:projected_power}
\end{figure*}

Among all possible interaction types, \textit{conflicting SS} is the most dynamic. The real action of the subject with a \textit{Soft} goal varies from giving up the assigned goal too easily or exerting too much effort. Hence, a number of \textit{conflicting SS} interactions might resemble similar patterns as in the rest of the interaction types. However, \textit{conflicting SS} contains unique situations where subjects take more turns to resolve the conflict assuming the partner can change their mind. Fig. \ref{fig:power_by_interaction_types}d shows an example where both partners initially opposed each other causing significant bifurcation, but subject 2 (red) decides to concede afterward.

% challenge: subjects do not act differently upon assigned goals. Different level of tactile sensitivity: follower person might be perceived as Soft goal.. etc

\subsection{Action identification}

The main challenge in an application like ours is detecting individual actions in a time-dependent signal. To detect push-pull actions we exploit patterns in the power signal. When an agent exerts significant force causing the object to move it creates peak shapes in power signals. In this work, we devised an action detection algorithm in a similar fashion as in \cite{rysbek_physical_action2021}. Specifically, we focus on the rising edge of the peak in the power signal. A peak associated with the action is determined by the spatiotemporal filter. As a temporal constraint, human reaction time for touch stimulus (0.15s) \cite{murchison1934handbook} and simple reaction time (0.25s) \cite{woodsFactorsInfluencingLatency2015} were considered. 

An example of action detection is shown in Fig. \ref{fig:projected_power}. The algorithm identified action regions shaded in blue and red for subjects 1 and 2. In the power plot (bottom left), subject 2 made a significant effort to cause the object to move. Subject 2 (blue) showed little opposition with a short delay. After the shaded area, tension decreases and follows to the execution phase. When the power in the neutral direction is seen in a different goal direction, it is clear that actions behind this movement were intended in towards goal 3 (Fig. \ref{fig:projected_power} lower right). Although subject 1 managed to apply opposing force which did not cause a motion, the power signal in goals 1 and 2 remained low. Thus, individual actions and their intended goal directions are interpreted.

% % Based on the observations, we hypothesized that the first dominant hump in the handle power describes the first intent of an agent. To confirm our assumption, we implemented an algorithm that identifies the beginning and end of these initial actions for each agent. 

% % We input the powers applied in each handle and the algorithm first finds the initial peak where its value is above a certain threshold and has sufficient width (enough duration of interaction). 

% The minimum width is determined by the average simple human reaction time, approximately 0.25 secs~\cite{woodsFactorsInfluencingLatency2015}. We took the $8\%$ of the local maxima power and find the intersection time, and set it as the rising moment. The agent showed an intent toward a certain goal if the peak power is positive and opposed to an intent if the peak is negative. 

% If both rising moments of the dyads occurred within the average human reaction time for touch stimulus, approximately 0.15 secs~\cite{murchison1934handbook}\cite{gescheidertactile2008}, both actions happened simultaneously. Otherwise, the one with a faster rising moment leads the initial state and its partner reacts to the movement. 
% The moments after the peak illustrates the next state, whether it goes to collaboration mode or to the next round of interaction, and thus we determined the local maxima time as the end of the initial state. 

As mentioned in Section \ref{sec:Deliberate}, force components due to grasping and walking interferes with the direction of pull and create ambiguity in identifying the intended goal direction. That is, maximum power among projected goal directions is not necessarily the intended goal. This problem can be solved as a classification problem by aggregating features extracted from force, velocity, and power signals from the action phase. Fig. \ref{fig:lda} demonstrates the clusterability of the intended direction for all interactions involving \emph{hard} goal 3. Table \ref{tbl:lda_eval}, reports the clustering score for H1, H2, H3 cases. Therefore, we show the feasibility of interpreting symbolic events from the force-kinematic signals.

% We applied Linear Discriminant Analysis (LDA) based on multiple measurements projected from subjects with \textit{Hard} goals to each goal location. Fig.~\ref{fig:lda} shows a sample of the subjects with \textit{Hard} goal 3 (H3). We used Calinski-Harabasz (CH), Davies–Bouldin (DB), and Silhouette (S) scores to evaluate whether the projected measurements are reasonably separable from the given goal direction. Table.~\ref{tbl:lda_eval} illustrates the evaluation metric scores for each \textit{Hard} goal case.

% In most cases identifying the intended direction is clear. But, in some cases where grasp force/forces due to motion has adversary effect, intent identification is not clear. This is segway to figure 8.
    
% First mover
%  - correlation with model and assigned goal intent
%  - Interaction force in Ehsan's PM models the grasping force

% How humans conduct such negotiations.

% 5N sensable range

% What does the first bumps tell
% strong intent, weak intent
% some cases we can predict the outcome.
% if intent is strong enough follows to collaboration

% - Check the accuracy of intent with assigned goals.
% - True intent signal is hidden under the noise of grasping force, and locomotion (gait patterns)

% semi conscious interacton

% acting upon assigned goal

% \textcolor{red}{we different ways out of conflict compared to turkish group, where conflict is only avoided by submitting to the human partners.}

%% file: Conclusion.tex
\section{Conclusion}
\label{sec:conclusion}

This work attempts to address an important problem in human-robot co-manipulation where dyads actively communicate with force signals. The task is grounded on a realistic world scenario where participants decide to move an object collaboratively to an uncertain goal. A taxonomy is presented that captures the diversity of human behavior conditioned on how much each participant possesses task-related information and extends human-human interaction scenarios studied in the literature. Based on that the human study experiment was designed. We emphasize that interactions that involve unknown goals are the least studied in robotics research. Therefore, we skewed the distribution of corresponding cases in the experiment. Empirical analysis of negotiation time shows the statistical significance of different categories of the taxonomy.

Furthermore, we observed due to the delay in reaction, the participants seem to communicate and express intent in epochs. To identify such human intents we propose novel features derived from force-velocity signals. Presented features are designed to accommodate arbitrary locations and numbers of goals, crucial for robot functionality in unstructured and changing environments. Additionally, projected features allow the mapping of low-level signals into symbolic action -- input to a high-level reasoning module. A preliminary analysis of proposed features demonstrates discriminating capacity where the corresponding feature resonates consistently when a participant shows intent.

The overreaching goal of this research is to develop a high-level robot controller that can physically collaborate with humans by taking the human lead as well as taking initiative when needed. This necessitates 3 major components of the controller: (1) the ability to understand human intent, (2) clearly express its own intent and (3) action policy that reaches the consensus. The clustering analysis presented in this paper demonstrates the feasibility of training a classifier to recognize human intent in real-time (component 1). The future continuation of this research involves designing a real-time human intent classifier and high-level robot controller with human intent feedback.

The ability of humans to engage in a vast diversity of collaborative tasks sets high expectations for robots to be used in daily life. Robots that lack human interactive controllers have much less use in practice. Therefore, the authors believe this work contributes to filling the gap in the literature to attain more fluid human-robot interaction. The application of such systems is vital in practice ranging from humanoids, assistive robots, rehabilitation, and human-in-the-loop automation.